# HIERARCHICAL DIGITAL IMAGE INPAINTING USING WAVELETS

S.Padmavathi [1], B.Priyalakshmi,[2] Dr.K.P.Soman[3]

[1]Department of Information Technology,
Amrita Vishwa Vidyapeetham, Coimbatore, India
`s_padmavathi@cb.amrita.edu`
[2]Department of Information Technology,
Amrita Vishwa Vidyapeetham, Coimbatore, India
`priyadeepu2@gmail.com`
[3]Center of Computational Engineering & Networking,
Amrita Vishwa Vidyapeetham, Coimbatore, India
`kp_soman@amrita.edu`

*ABSTRACT*

*Inpainting is the technique of reconstructing unknown or damaged portions of an image in a visually plausible way. Inpainting algorithm automatically fills the damaged region in an image using the information available in undamaged region. Propagation of structure and texture information becomes a challenge as the size of damaged area increases. In this paper, a hierarchical inpainting algorithm using wavelets is proposed. The hierarchical method tries to keep the mask size smaller while wavelets help in handling the high pass structure information and low pass texture information separately. The performance of the proposed algorithm is tested using different factors. The results of our algorithm are compared with existing methods such as interpolation, diffusion and exemplar techniques.*

## 1. INTRODUCTION

The objective of image inpainting or retouching, is to reconstruct missing or damaged portions of image in an unnoticeable way, producing repaired images with satisfactory visual quality. Majority of the Image Inpainting Algorithms achieve such a task by using Partial Differential Equations. The applications of inpainting includes reconstruction of small damaged paintings, removal of superimposed text, removal of scratches due to folding of images, removal of an object in an image etc.

Many graphics editing softwares have incorporated a tool for inpainting. The difference between these tools and the inpainting algorithm which we follow is that, the former requires the user to select both the region to be inpainted and the pattern that is to be filled in the unknown region. Hence it creates overhead on the user as he/she may not be sure of the exact pattern to be used to fill while the latter takes only the region to be inpainted as input from the user. The result of the tools will have blocky effects when it just replaces the original region with the new pattern. Moreover the process becomes tedious when the user has to fill large number of smaller areas.

The process of inpainting, as mentioned in the Fig.1 includes masking out the unknown region selected by the user, and the inpainting technique is applied to fill the masked region. Many inpainting techniques have been proposed which retouches the image in an effective manner. Some of the existing techniques are interpolation, isophotic diffusion and exemplar based inpainting. Interpolation and diffusion based techniques work well for smaller areas but fails to reproduce texture properly. It also results in blurring of edges. Whereas exemplar based method works well for larger areas but fails in proper reproduction of definite shapes. It results in excessive propagation of texture and hence damaging the larger structures in the image. In this paper, the structure and texture information are separated and the coarser structures are handled first and then moving to finer details. Multi resolution property of the wavelets makes it desirable to be used in the process of inpainting. As the wavelets has the property of separating the low pass and high pass coefficients, it provides us the structure and texture information of the image. Inpainting algorithm is applied to four subbands of the image formed after applying Wavelet Transform. This gives a better reconstruction of the images.

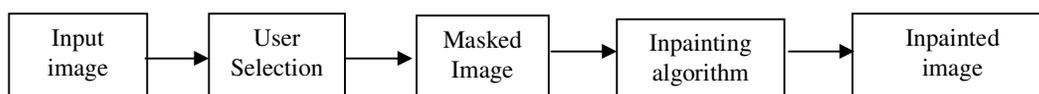

Figure 1. Inpainting Process

## 2. RELATED WORK

Interpolation methods are the primitive methods which can be used for inpainting. In the interpolation method[9], the neighboring pixels are considered for filling the inpainting area. The masked pixels are replaced with the the average of the neighboring pixels. The technique gives better result for the uniform area and fails in high structured regions.

Bertalmio et al. [1] pioneered a digital image inpainting algorithm based on partial differential equations (PDEs). Bertalmio et al., described an algorithm where the isophotes are extended or prolonged inwards from the boundary of mask region. The structure of the image on the boundary of mask region is extended inward. Anisotropic diffusion[6] is applied where the gradient vector is computed and rotated by $\pi/2$ radians to obtain the direction of the isophote lines. Though the algorithm works well for small textured images, it fails in large textured images. Other methods involving partial differential equation and concentrating on smaller structures could be found in [2],[4] and [5].

Texture synthesis [10],[13] is another way of synthesizing the missing area. Texture synthesis based inpainting could be extensively found in [3],[7],[8],[11], and [12]. Criminsi et al., [7] have presented exemplar based technique for filling image regions. It is derived from patch based texture synthesis. In their algorithm, both structure and texture information is propagated into the mask region. The algorithm works by taking a patch around a pixel on the boundary and replacing it with the best patch found by searching in the source region. The comparison of structure based and texture based inpainting is available in [14]. Hierarchical TV inpainting is discussed in [15].

## 3. PROPOSED METHOD

The inpainting problem is depicted in Figure 2 to illustrate the notations involved in image inpainting. Initially, the algorithm allows the user to select target region, Ω which is to be inpainted. The region to be inpainted is also called as mask which is represented as gray area in the figure. The region surrounding Ω which forms a boundary between the target region and other region is denoted as ∂Ω. The Source region, Φ is the region of the image which is not to be inpainted. This represents the white portion of the image except the masked region, Ω. The inpainting algorithm fills Ω starting from ∂Ω using the information in Φ. It updates ∂Ω while filling which in turn shrinks Ω. The exemplar method discussed in [7] uses patches on ∂Ω involving some unknown pixels of Ω and some known pixel of Φ as shown in Figure 3. For a pixel 'p' on the boundary, the patch $\psi_p$ is formed with 'p' as the centre. In the figure inner red square represents the pixel on the boundary and the outer green square represents the patch. A patch that is similar to the known pixels is searched in the entire image, (i.e.) patch that closely matches with the known pixels of $\psi_p$ is searched in Φ. The patch that yields minimum SSD (Sum of Squared differences) value is taken as the best match. This is called as the exemplar patch, $\psi_q$. The values corresponding to the unknown pixels are copied from the best matched patch. The unknown pixels of $\psi_p$ are copied from corresponding locations of $\psi_q$.

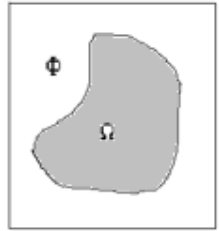  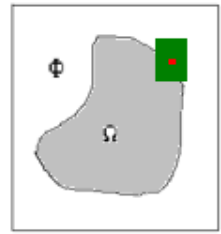

Figure 2 Inpainting Problem          Figure 3 Patch based exemplar method

A higher number of known pixels in the patch $\psi_p$ increase the confidence of accuracy. If the pixel 'p' lies on an edge touching the ∂Ω, copying the pixels leads to copying of edges and hence avoids blurring of edges. If any other patch near 'p' is considered first for restoration, edges(structure) will not be extended correctly. Hence the pixels on the boundary and the pixels near the edges have to be given a higher priority for restoration process. These priorities are termed as Confidence and Structure terms respectively. The patch priority is computed as the product of confidence term and Structure term. The patch with maximum priority value is restored first. Once when the highest priority patch is filled, there will be a change in the confidence values and the boundary. The confidence term of that particular patch will be updated as the sum of all the confidence values of the newly filled pixels divided by the total number of pixels in the patch. The new boundary is detected and the process is repeated until all the patches are filled and till the number of boundary pixels becomes zero. In this paper, the exemplar based method is adopted in a hierarchical multi resolution approach. Filling the coarser details first and then the finer details tend to improve the accuracy. Moreover handling the structures separately ensures the structure continuity in the image.

Wavelet transforms are well known for their multi resolution property. The scaled and translated basis functions of wavelet transform [16] are given in Eqn. (1) and (2). For an image 'f' of size M x N, the wavelet coefficients at any particular level 'j', can be obtained through Eqn. (3) and (4). Eqn. (3) gives the approximation coefficients and Eqn. (4) gives the detail coefficients.

$$g_{j,m,n}(x, y) = 2^{j/2} \cdot g(2^j x - m, \; 2^j y - n) \qquad (1)$$

$$h^i_{j,m,n}(x, y) = 2^{j/2}.h(2^j x - m, 2^j y - n), \qquad i = \{HL, LH, HH\} \qquad (2)$$

$$W_g(j_0, m, n) = \frac{1}{\sqrt{MN}} \sum_{x=0}^{M-1} \sum_{y=0}^{N-1} f(x,y) g_{j_0,m,n}(x,y) \qquad (3)$$

$$W^i_h(j,m,n) = \frac{1}{\sqrt{MN}} \sum_{x=0}^{M-1} \sum_{y=0}^{N-1} f(x,y) h^i_{j,m,n}(x,y), \qquad i = \{HL, LH, HH\} \qquad (4)$$

Hence applying discrete wavelet transform (DWT) to an image decomposes the image in four sub bands LL, HL, LH and HH of reduced spatial resolution as shown in figure 4. Here the LL corresponds to the Low pass band (approximation coefficients) and majorly contains the texture information. The remaining three bands correspond to the high pass bands (detail coefficients) and contain the structure information. The resolution of the LL band could be further reduced by repeatedly applying the wavelet transform. At each level certain amount of fine details are shed off the LL band leaving the coarser details. A wavelet decomposed image after applying wavelet transform thrice is shown in figure 5. At level 3 the LL band contains the coarser texture information and HL,LH and HH contains the coarser structure information. Inpainting at the lowest resolution ensures the filling of coarser details first. Inpainting in each sub band separately ensures the handling of texture and structure information separately. Once all the sub bands in a particular level is filled they are combined to form the LL band of the next higher resolution through inverse wavelet transform as specified in Eqn. (5)

$$f(x,y) = \frac{1}{\sqrt{MN}} \sum_m \sum_n W_g(j_0,m,n) g_{j_0,m,n}(x,y) + \frac{1}{\sqrt{MN}} \sum_i \sum_{j}^{\infty} \sum_m \sum_n W^i_h(j,m,n) h^i_{j,m,n}(x,y) \qquad (5)$$

The process is repeated in the higher resolution levels until the original image size.

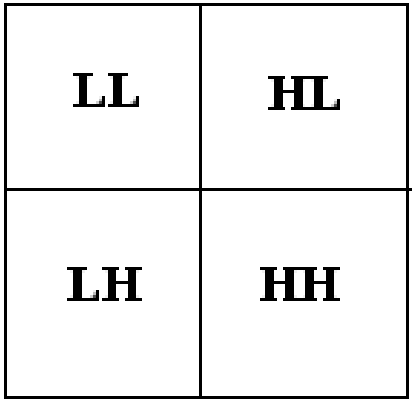 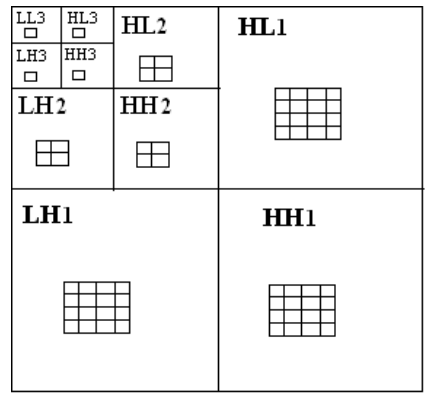

Figure 4: Image showing subbands after applying DWT

Figure 5: Image decomposition after 3 levels of DWT

The steps involved in the proposed algorithm are as follows
1. Select the area to be inpainted, the mask
2. If mask size >Threshold T

    a. Decompose the image/LL subband using wavelet transform as in equation (3) and (4)
    b. Identify the mask pixels called target pixels in the decomposed image.

3. Compute the confidence term for all target pixels

Initially the confidence value is assigned as 0 for the target region($\Omega$) and 1 for the source region($\Phi$). The confidence value for a patch $\psi_p$ is calculated as in equation (6)

$$C(p) = \frac{\sum_{q \in \Phi \cap \Psi_p} C(q)}{|\Psi_p|} \qquad (6)$$

Where, C(q) is the confidence value of those pixels belonging to the source region and the patch $\psi_p$. The denominator is the cardinality of the patch.

4. Compute the structure term for all target pixels

The structure term ensures the continuity of the structure in an image. The structure term for any high pass band is calculated as in equation (7). The structure term for the LL band is calculated as the average value of the three high pass bands.

$$S(p) = \frac{\sum_{q \in \Phi \cap \Psi_p} e(q)}{|\Psi_p|} \qquad (7)$$

Where $e(q)$ specifies the coefficient values of those pixels belonging to the source region and the patch $\psi_p$

5. Compute patch priorities $P(p)$ as in equation (8)

$$P(p) = C(p) * S(p), \qquad (8)$$

where C(p) is the confidence term and S(p) is the structure term.

6. Find the patch $\Psi p$ with the maximum priority as in equation (9),

$$\textit{i.e., } p = \arg\max p \textit{ for all } P(p) \qquad (9)$$

7. Find the best matching patch $\Psi q \in \Phi$ in source region that minimizes d($\Psi p; \Psi q$), where d is the Sum of Squared Distance(SSD).
8. Copy image data from $\Psi q$ to $\Psi p$ for all pixels belonging to the target region.
9. Update $C(p)$ and the S(p) for the newly filled pixels.
10. . Repeat steps 3 to 9 until all target pixels in each sub band in the current level are filled.
11. .Reconstruction to next level
   Apply inverse wavelet transform as in equation (5) and construct the LL band of the next higher resolution level from the four sub bands of the current resolution.
12. Mapping of mask pixels

As the target pixels in the lower resolution level are filled, after reconstruction few target pixels in the higher level will also be filled. Mapping between the pixels in the higher and lower levels has to be made before filling the current level.

13. Steps 3 to 12 are continued until the original image is inpainted.

## 4. EXPERIMENTAL RESULTS

Experiments have been conducted for various images with variable mask size and shape. The mask is chosen on uniform areas, high contrast areas etc. The image is decomposed using Haar wavelets and the sub bands in the lowest resolution level are filled as explained in the previous section. Interpolation and diffusion are some of the existing techniques which work well for smaller mask sizes and exemplar based methods perform better for larger mask sizes. The proposed method is compared with these existing methods. Interpolation and diffusion techniques outperform other methods in speed when the mask is chosen from a uniform area. They also perform well for large number of smaller masks. However their performance decays as the mask size increases. They fail drastically for textured images. The results for a textured low contrast and high contrast images are shown in Figure 6 and 7 respectively. In each set (a) shows the image with the mask area made zero, (b) shows the result of diffusion method, (c) shows the result of nearest neighbor interpolation method and (d) shows the result of the proposed method.

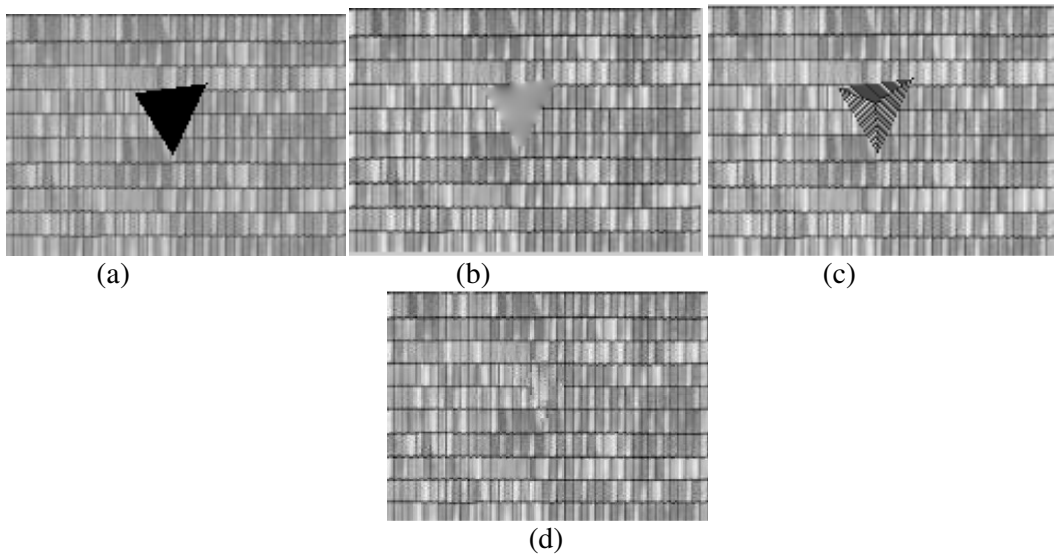

(a)　　　　　　　　　　　(b)　　　　　　　　　　　(c)

(d)

Figure 6. (a) Image with masked area, (b) Result of diffusion technique, (c) Result of nearest neighbour interpolation technique, (d) Result of the Hierarchical DWT method

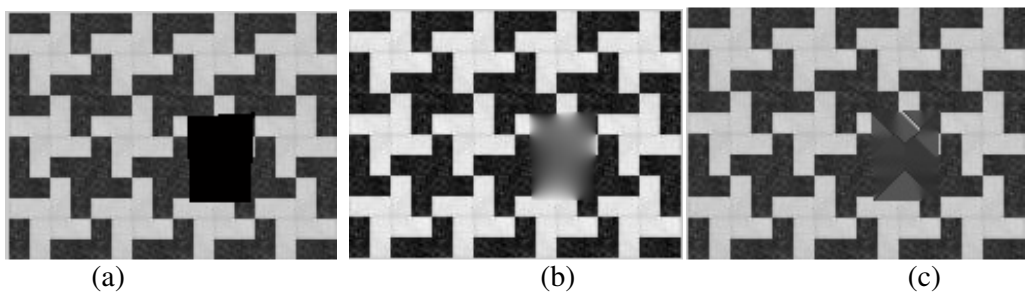

(a)　　　　　　　　　　　(b)　　　　　　　　　　　(c)

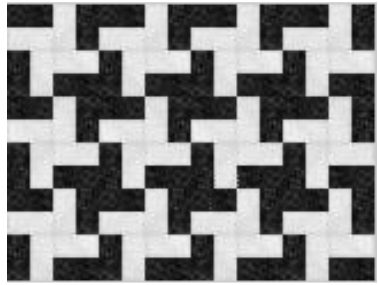

(d)

Figure 7. (a) Image with masked area, (b) Result of diffusion technique, (c) Result of nearest neighbour interpolation technique, (d) Result of the Hierarchical DWT method

Exemplar methods perform well for textured images, but do not propagate the structure in natural scenic images as shown in Figure 8.(b) whereas the hierarchical wavelet method performs better as shown in Figure 8(c).

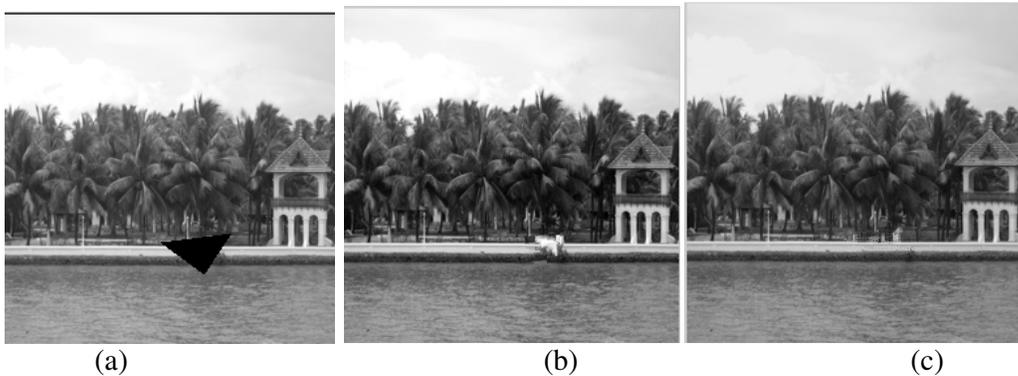

(a)                                              (b)                                              (c)

Figure 8. (a) Image with masked area, (b) Result of Exemplar method, (c) Result of Hierarchical DWT

From the results it could be seen that the hierarchical wavelet based inpainting performs better than the existing methods. The selection of the decomposition levels plays a crucial role in the quality of reconstruction. If a smaller value is chosen the structure propagation gets affected and if a larger value is chosen it introduces minor distortions in the reconstructed area as shown in red circle in Figure 9(b). Figure 9(a) shows the result of proper reconstruction at lower level. Since the coefficients in the transformed domain are copied into the inpainting area, the reconstructed image has varied brightness and contrast, which is proportional to the decomposition level. However the decomposition level for proper reconstruction is found to be proportional to the size of inpainting area.

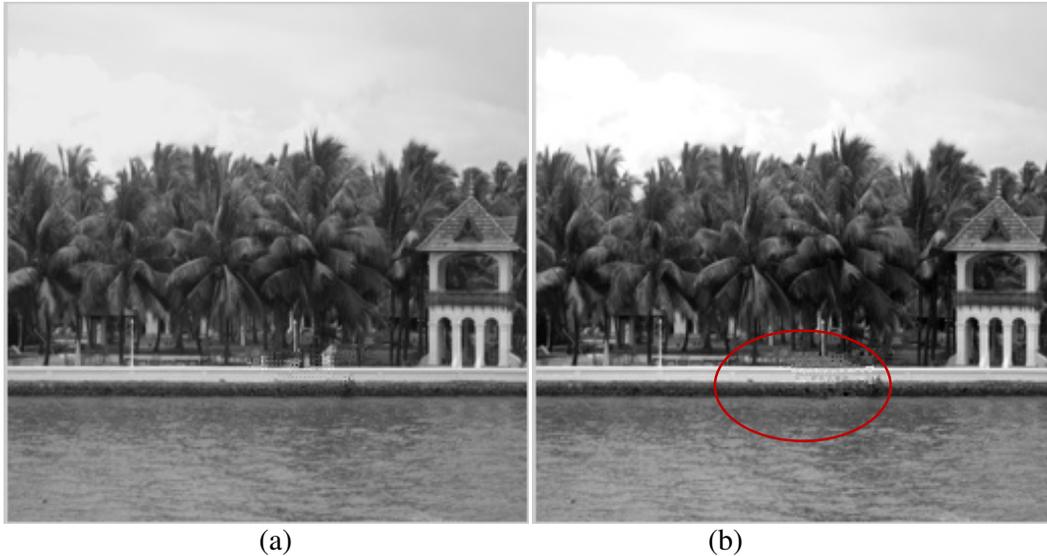

(a)            (b)

Figure 9 (a) Image inpainted from appropriate level (b) Image inpainted with excessive level, Inpainted area is marked in red circle

## 5. CONCLUSION

Digital image inpainting offers a digital technique for restoring a damaged image. The algorithm requires the user to specify the damaged portion manually. It generates the damaged portions using other portions of the same image. It cannot generate a portion which is not available in the undamaged portions. Majority of the algorithms concentrate on images with smaller damaged portions. The quality of performance drops as the mask size increases. Exemplar based methods perform well for larger mask size but fails in larger structure propagation. Hierarchical method proposed in this paper tries to utilize the advantage of the Exemplar based method by handling the structures separately through wavelets. The structure propagation is better when inpainted from the lower resolution level. Selection of the decomposition level depends on the mask size. Though this method produces visual quality better than other methods it changes the overall brightness and contrast of the image as the number of levels increases. The inpainting effort increases as the algorithm is applied to all the sub bands at various levels.

[6]  P. PERONA AND J. MALIK, "Scale-Space Edge Detection Using Anisotropic Diffusion" IEEE Transactions on Pattern Analysis and Machine Intelligence, Vol. 12, No.7, July 1990.

[7]  A. CRIMINISI, P. PÉRES AND K. TOYAMA, "Object Removal by Exemplar-Based Inpainting" Proceedings of the 2003 IEEE Computer Society Conference on Computer Vision and Pattern Recognition (CVPR'03).

[8]  H. IGEHY AND L. PEREIRA, "Image Replacement through Texture Synthesis" Proceedings of the IEEE International Conference on Image Processing, October 1997.

[9]  A.C. KOKARAM, R.D. MORRIS, W.J. FITZGERALD AND P.J.W. RAYNER, "Interpolation of Missing Data in Image Sequences" IEEE Transactions on Image Processing. Vol. 4. no.11, Nov. 1995, pp 1509-1519.

[10] A. EFROS ANDW.T. FREEMAN,." Image quilting for texture synthesis and transfer", In Proc. ACM Conf. Comp. Graphics (SIGGRAPH), pages 341–346,Eugene Fiume, August 2001.

[11] A. HERTZMANN, C. JACOBS, N. OLIVER, B. CURLESS, AND D. SALESIN,." Image analogies",. In Proc. ACM Conf. Comp. Graphics (SIGGRAPH), Eugene Fiume, August 2001.

[12] S. RANE, G. SAPIRO, AND M. BERTALMIO., "Structure and texture filling-in of missing image blocks in wireless transmission and compression applications". In IEEE. Trans. Image Processing, 2002.

[13] L. LIANG, C. LIU, Y.-Q. XU, B. GUO, AND H.-Y. SHUM. "Real-time texture synthesis by patch-based sampling", ACM Transactions on Graphics, 2001.

[14] S.PADMAVATHI, K.P.SOMAN, "Comparative Analysis of Structure and Texture based Image Inpainting Techniques", International Journal of Electronics and Computer Science Engineering,( IJECSE )Volume 1, Number 3, June 2012, pp1062-1069. http://www.ijecse.org/wp-content/uploads/2012/06/Volume-1Number-3PP-1062-1069.pdf

[15] S.PADMAVATHI, N.ARCHANA, K.P.SOMAN, "Hierarchical Approach for Total Variation Digital Image Inpainting",  International Journal of Computer Science, Engineering and Applications (IJCSEA), Volume 2, Number 3, June 2012, http://airccse.org/journal/ijcsea/papers/2312ijcsea16.pdf

[16] R.C. GONZALES AND R.E. WOODS, "Digital Image Processing", Second Edition, Prentice Hall, Inc. 2002.